\documentclass[conference]{IEEEtran}
\IEEEoverridecommandlockouts
\usepackage{cite}
\usepackage{amsmath,amssymb,amsfonts}
\usepackage{algorithmic}
\usepackage{graphicx}
\usepackage{textcomp}
\usepackage{xcolor}
\usepackage{booktabs}
\usepackage{multirow}
\def\BibTeX{{\rm B\kern-.05em{\sc i\kern-.025em b}\kern-.08em
    T\kern-.1667em\lower.7ex\hbox{E}\kern-.125emX}}

\begin{document}

\title{GPT-3 Powered Information Extraction for Building Robust Knowledge Bases}

\author{
\IEEEauthorblockN{Ritabrata Roy Choudhury}
    \IEEEauthorblockA{
        \textit{School of Computer Engineering} \\
        \textit{Kalinga Institute of Industrial Technology} \\
        Bhubaneswar, Odisha, India \\
        \textit{email:ritabrata2003.rrc@gmail.com}
    }
    
    \and
     \IEEEauthorblockN{Soumik Dey}
    \IEEEauthorblockA{
        \textit{School of Computer Engineering} \\
        \textit{Kalinga Institute of Industrial Technology} \\
        Bhubaneswar, Odisha, India \\
        \textit{email:soumikdey2@gmail.com}
    }
}

\maketitle

\begin{abstract}
This work uses the state-of-the-art language model GPT-3 to offer a novel method of information extraction for knowledge base development. The suggested method attempts to solve the difficulties associated with obtaining relevant entities and relationships from unstructured text in order to extract structured information. We conduct experiments on a huge corpus of text from diverse fields to assess the performance of our suggested technique. The evaluation measures, which are frequently employed in information extraction tasks, include precision, recall, and F1-score. The findings demonstrate that GPT-3 can be used to efficiently and accurately extract pertinent and correct information from text, hence increasing the precision and productivity of knowledge base creation. We also assess how well our suggested approach performs in comparison to the most advanced information extraction techniques already in use. The findings show that by utilising only a small number of instances in in-context learning, our suggested strategy yields competitive outcomes with notable savings in terms of data annotation and engineering expense. Additionally, we use our proposed method to retrieve Biomedical information, demonstrating its practicality in a real-world setting. All things considered, our suggested method offers a viable way to overcome the difficulties involved in obtaining structured data from unstructured text in order to create knowledge bases. It can greatly increase the precision and effectiveness of information extraction, which is necessary for many applications including chatbots, recommendation engines, and question-answering systems.
\end{abstract}


\section{Introduction}
With the rapid increase in the generation of  biomedical research and clinical text , it has become more and more essential for both researchers and practitioners to convert large amounts of biomedical text into structured data. Recently, pre-trained language models (PLMs), which can be either general-purpose or specialized for biomedicine, have significantly improved the ability to extract information from the biomedical text in various tasks \cite{lee2020biobert}.

The development of Generative Pre-trained Transformer, GPT-3 \cite{brown2020language}, a new pre-trained language model, represents a significant advancement in the field of natural language processing. Unlike previous models which required extensive fine-tuning for specific tasks, GPT-3 can generalize unseen cases after being provided with just a few in-context examples. This opens up many new possibilities for NLP systems, including expanding emails, entity extraction from text, and generating code based on natural language instructions with only a few demonstration examples.

Newly released pre-trained language models (PLMs), including GPT-3, Megatron-Turing NLG \cite{smith2022using}, and the Switch Transformer\cite{fedus2021switch}, have many thousands parameters and have demonstrated remarkable achievement in natural language processing (NLP) tasks using a new research paradigm called ``in-context learning." PLMs can utilize their natural language generation skills to complete prompts or pieces of text in a manner similar to how humans approach a given task. By utilizing in-context learning, these large models can tackle various NLP issues without requiring updates to their parameters. This approach results in significant savings in terms of data annotation and engineering costs compared to traditional model training methods. It is worth noting that GPT-3's in-context learning produces competitive outcomes in numerous NLP tasks, even when supplied with only a limited number of demonstrative examples in the prompt.

As there are multiple potential applications for biomedical information extraction and the cost of biomedical annotations is high, alongside the challenges in model training, in-context learning has become an appealing option for biomedical use cases. To test the feasibility of this approach, we conducted a thorough and systematic study to compare the effectiveness of GPT-3 in-context learning and BERT-sized PLM fine-tuning  \cite{devlin2018bert} in the few-shot setting for named entity recognition (NER) and relation extraction (RE) – both important biomedical information extraction tasks. In order to maintain consistency and comprehensiveness, we utilized all the biomedical NER and RE tasks accessible in the BLURB benchmark \cite{gu2021domain}. To ensure an accurate evaluation, we adopted the true few-shot setting introduced by \cite{perez2021true} to avoid overestimating the models' few-shot performance through model selection on a large validation set.

The main contributions of this paper are the following: 
\begin{enumerate}

    \item The paper proposes an in-context learning approach for Information Extraction in Knowledge Base Construction using GPT-3, which involves creating a structured prompt \cite{perez2021true}, utilizing a k-nearest neighbor module \cite{agirre2022proceedings}, and integrating logit biases and contextual calibration for NER and RE. The approach is evaluated using biomedical NER and RE tasks, and the results show its potential for practical use cases compared to fine-tuned models with no additional cost.

    \item The proposed model presents a unique approach to the implementation of information obtained through the aforementioned extraction processes. This approach involves the organization and construction of a knowledge base in a structured and systematic manner. The model's methodology is distinct from conventional methods, and it seeks to optimize the efficiency and accuracy of the information extraction process by utilizing an innovative approach to knowledge base construction.
 
    
\end{enumerate}

Overall, the proposed approach's novelty lies in adapting GPT-3 to the task of Information Extraction for KBC, its ability to understand the context and meaning in natural language text, and the demonstration of its potential in a real-world setting. The research opens up new avenues for using advanced natural language processing techniques to extract structured information from unstructured text and has the potential to significantly improve the efficiency and accuracy of knowledge base construction.

The resultant knowledge base may be used to support a range of applications, including chatbots, intelligent search engines, and recommendation systems. It can also be utilized in a variety of sectors, including e-commerce, healthcare, and finance. KBC is an important field of study and development in artificial intelligence and natural language processing since the accuracy and quality of the knowledge base are essential to the functionality and success of these applications.
\section{Related Work}

In a variety of tasks, such as text classification, natural language inference, machine translation, question answering, table-to-text generation, and semantic parsing, GPT-3 in-context learning \cite{brown2020language} has been found to be competitive \cite{liu2019roberta,shin2021constrained} against supervised baselines \cite{brown2020language,zhao2021calibrate}. Many methods have been developed to improve its performance, such as removing biases through calibration \cite{zhao2021calibrate,malkin2022coherence}; optimizing prompt retrieval \cite{liu2019roberta,wolf2020transformers,shin2021constrained}; prompt ordering \cite{liu2019roberta}; and optimizing prompt design\cite{perez2021true}. Much research hasn't been done on how well GPT-3 performs while learning in context for information extraction tasks. Smaller GPT-3 models are assessed by \cite{zhao2021calibrate} using a modified slot-filling task where every sample has at least one item of interest. Moreover, \cite{epure2021probing} assess the GPT-2's in-context learning performance using open-domain NER datasets that have been altered to maintain a particular ratio of empty to non-empty cases. Our biological NER prompt design closely references both of these pieces. We believe that our work is among the first to thoroughly assess the in-context learning capabilities of GPT-3 on IE tasks.

Several additional research paths examine ways to reframe NLP problems as language generation challenges aside from the work on in-context learning. In order to enhance few-shot learning in smaller pre-trained language models, \cite{schick2020s} reformulated the text classification and natural language inference tasks using a variety of manually created cloze-style templates as prompts.\cite{gu2021domain} investigate a comparable environment but make use of an external language model to produce such templates. Both of these highlight the value of utilizing different prompt designs. In the same vein, \cite{cabot2021rebel}  reformulate relation extraction benchmarks as an end-to-end sequence-to-sequence work to attain state-of-the-art performance.\cite{raffel2020exploring} introduced the multi-task sequence-to-sequence paradigm, and several works in the biomedical domain \cite{raval2021exploring,phan2021scifive,parmar2022boxbart} follow it. These works outperform previous methods on many tasks, including side effect extraction, NER, RE, natural language inference, and question answering. Several of these initiatives to rephrase IE tasks as sequence-to-sequence activities served as major inspirations for our quick design.

According to \cite{perez2021true} past research overestimates PLMs' few-shot learning capabilities by choosing models and prompts from huge validation sets. Several research in this approach has used this setting in an effort to estimate few-shot performance more precisely \cite{logan2021cutting,schick2022true,liu2019roberta}.

Using the GPT-3 API directly results in a subpar performance in the biomedical domain, according to previous research analyzing GPT-3's in-context learning capabilities on biomedical NLP tasks \cite{moradi2021gpt}. They provide experimental findings on five biomedical NLP datasets for various tasks, such as connection extraction. We intend to deliver a thorough and in-depth evaluation of biomedical IE in our study by utilizing a well-established multi-dataset biomedical NLP benchmark and cutting-edge in-context learning methodologies. This will allow us to achieve the best performance to the best of our knowledge and capacity. However, the inadequacy of GPT-3 in-context learning for biological IE tasks is finally supported by our results and cannot be easily remedied with current methods. GPT-3 performs well on a different set of clinical Questions, according to a parallel study \cite{agrawal2022large}, which is an interesting task, such as a clinical one on the extraction of biological evidence. The reason for this unexpected difference in IE performance across the clinical and biomedical domains for in-context learning has to be investigated further.

\section{Proposed Approach}
In the following section, we present the two approaches we explored for named entity recognition (NER) and relation extraction (RE) utilizing the genuine few-shot scenario: adapting pre-trained language models (PLM) of BERT size and in-context learning of GPT-3. The proposed model's conceptual foundation draws from the seminal work of \cite{gutierrez2022thinking}.

\subsection{True Few-Shot Settings}

Recent research shows concerns regarding the reliability of few-shot learning in large pre-trained language models (PLMs) such as GPT-3 and minor PLM fine-tuning. The selection of models and prompts can be influenced by large validation sets. To address these concerns and avoid misjudging the performance of pre-trained language models on a small training set, we adopt the true few-shot configuration proposed by \cite{perez2021true}. In this approach, the model preferences are systematically based on the small training set rather than a large validation set. We use cross-validation on a set of 100 training samples to choose the prompt structure, the number of few-shot instances for each question, and the fine-tuning evaluation metrics for our experiments.

\begin{figure}
\centerline{\includegraphics[width=\linewidth]{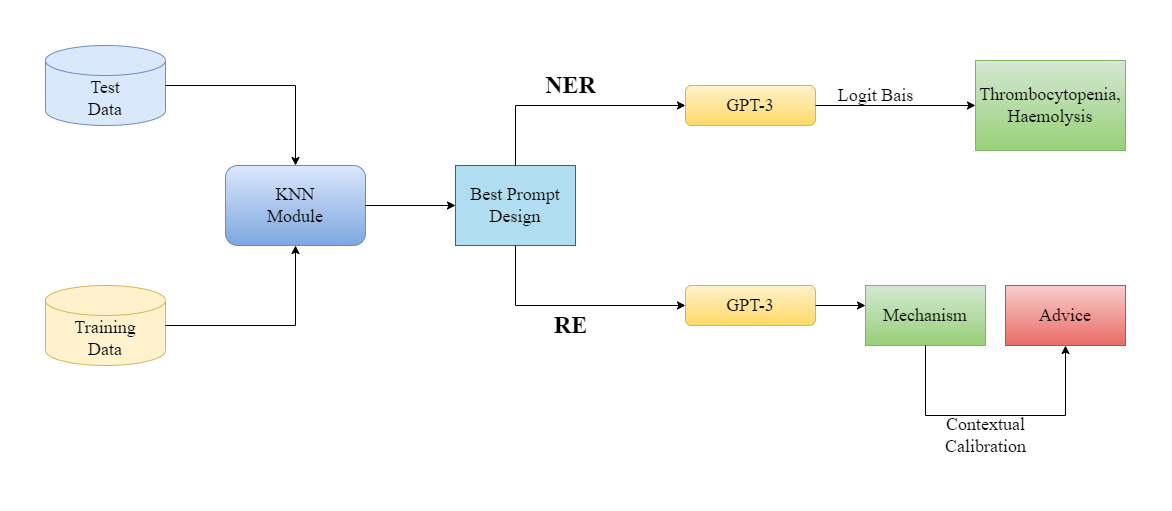}}
\caption{Representation of the proposed approach}
\label{fig:model}
\end{figure}

\subsection{Context-Based Learning in GPT-3}
The modifications we made to the named entity recognition (NER) and relation extraction (RE) objectives to accommodate in-context learning are described in this section. We also discuss other strategies we used to improve GPT-3's performance in biological IE through in-context learning, including our procedures for developing prompts and retrieving in-context examples..

\subsubsection{Sequencing of Tasks}  

To tackle NER and RE, we have transformed them into language generation tasks. We have employed a similar approach to other studies, which involves identifying all the item spans in the original phrase and merging them with a separator. Our method adds entities only once and trains the GPT-3 algorithm to generate a list of items separated by the chosen separator based on the input and its context.

To carry out relation extraction, we adopt the method proposed by \cite{cabot2021rebel} and utilize their technique of transforming each instance into a prompt, as shown in Figure \ref{fig:model}. Our prompt templates feature the original semantic representation of both the subject and object entities.

\subsubsection{Designing the Prompt}
The prompts must be properly chosen if you want the GPT-3 in-context learning to perform as well as possible (\cite{schick2022true}). We provide a methodical, task-independent method for creating GPT-3 prompts, which is divided into three key components : task assignment commands, a phrase introduction, and a recovery message. A label verbalizer is also defined for relation extraction to make it easier to produce natural language words that map relation categories. The ideal mix of prompt alternatives and the quantity of in-context examples contained in the prompt is determined using leave-one-out cross-validation (LOOCV), which we manually build a set of alternatives for each question section. We evaluate eight prompt options for each dataset to reduce expenses.

\subsubsection{Implementation of Logit Bias}
To avoid GPT-3 from generating tokens that were not present in the original text, we make use of the "logit bias" feature available in the OpenAI Completion API. This option limits the set of tokens that GPT-3 can generate by increasing the likelihood of a specific set of tokens. We increase the valuation each token in the initial text by 10, along with their chosen separator and the newline token, to be more accurate. After processing, we eliminate any forecasted entities which do not resemble any spans in the original phrase.

\par
We discovered that when we provided GPT-3 with a set of few-shot in-context examples, it became biased towards certain labels regardless of the test input. To address this issue, we used a technique in which the output is calibrated using a linear transformation that evenly distributes the label probabilities produced via GPT-3 when given one null prompt (where the test input is changed to "N/A").. They applied this approach to their RE task to remove context-induced biases.

\subsubsection{Dynamic Selection of in-Context Instances}
Improved performance in GPT-3 in-context learning can be achieved by dynamically selecting a small number of in-context instances for every test example. According to various research \cite{agirre2022proceedings}, this method entails choosing the most comparable examples from the training set to serve as a few-shot in-context prompt for every training sample using a k-nearest neighbor (kNN) retrieval module. RoBERTa-large has been chosen as the encoder for the kNN retrieval module since it has been found to be superior to other biomedical PLMs, sentence-transformer models \cite{reimers2019sentence}, and a BM25 baseline \cite{robertson2009probabilistic}. baseline.

\begin{table*}[h]
\centering
\caption{Comparison of Performance of Fine-tuned BERT-sized PLMs with GPT-3 In-Context}
\label{tab:comparison}
\begin{tabular}{|c|ccc|ccc|ccc|ccc|}
\hline
\multirow{2}{*}{} & \multicolumn{3}{c|}{\textbf{GPT-3 In-Context}} & \multicolumn{3}{c|}{\textbf{PubMedBERT-base}} & \multicolumn{3}{c|}{\textbf{RoBERTa-large}} & \multicolumn{3}{c|}{\textbf{BioBERT-large}} \\ \cline{2-13} 
                  & \multicolumn{1}{c|}{Precision} & \multicolumn{1}{c|}{Recall} & F1   & \multicolumn{1}{c|}{Precision} & \multicolumn{1}{c|}{Recall} & F1   & \multicolumn{1}{c|}{Precision} & \multicolumn{1}{c|}{Recall} & F1   & \multicolumn{1}{c|}{Precision} & \multicolumn{1}{c|}{Recall} & F1   \\ \hline
JNLPBA            & \multicolumn{1}{c|}{44.7}      & \multicolumn{1}{c|}{52.4}   & 48.3 & \multicolumn{1}{c|}{56.9}      & \multicolumn{1}{c|}{67.9}   & 61.9 & \multicolumn{1}{c|}{57.2}      & \multicolumn{1}{c|}{75.1}   & 65.0 & \multicolumn{1}{c|}{57.4}      & \multicolumn{1}{c|}{73.7}   & 64.6 \\ \hline
NCBI-disease      & \multicolumn{1}{c|}{55.2}      & \multicolumn{1}{c|}{49.0}   & 51.4 & \multicolumn{1}{c|}{68.5}      & \multicolumn{1}{c|}{67.6}   & 68.0 & \multicolumn{1}{c|}{64.3}      & \multicolumn{1}{c|}{68.7}   & 66.4 & \multicolumn{1}{c|}{59.6}      & \multicolumn{1}{c|}{67.0}   & 63.0 \\ \hline
BC5CDR-disease    & \multicolumn{1}{c|}{57.9}      & \multicolumn{1}{c|}{35.0}   & 43.6 & \multicolumn{1}{c|}{67.4}      & \multicolumn{1}{c|}{67.5}   & 67.4 & \multicolumn{1}{c|}{66.9}      & \multicolumn{1}{c|}{68.7}   & 67.7 & \multicolumn{1}{c|}{62.9}      & \multicolumn{1}{c|}{69.0}   & 65.8 \\ \hline
BC2GM             & \multicolumn{1}{c|}{43.0}      & \multicolumn{1}{c|}{40.8}   & 41.4 & \multicolumn{1}{c|}{55.4}      & \multicolumn{1}{c|}{57.9}   & 56.5 & \multicolumn{1}{c|}{49.7}      & \multicolumn{1}{c|}{56.3}   & 52.7 & \multicolumn{1}{c|}{53.6}      & \multicolumn{1}{c|}{59.2}   & 56.2 \\ \hline
BC5CDR-chem       & \multicolumn{1}{c|}{74.7}      & \multicolumn{1}{c|}{71.4}   & 73.0 & \multicolumn{1}{c|}{86.1}      & \multicolumn{1}{c|}{88.6}   & 87.3 & \multicolumn{1}{c|}{82.1}      & \multicolumn{1}{c|}{87.3}   & 84.6 & \multicolumn{1}{c|}{84.8}      & \multicolumn{1}{c|}{87.3}   & 86.0 \\ \hline
\textbf{NER Average} & \multicolumn{1}{c|}{55.1}      & \multicolumn{1}{c|}{49.7}   & 51.5 & \multicolumn{1}{c|}{66.9}      & \multicolumn{1}{c|}{69.9}   & 68.2 & \multicolumn{1}{c|}{64.0}      & \multicolumn{1}{c|}{71.2}   & 67.2 & \multicolumn{1}{c|}{63.7}      & \multicolumn{1}{c|}{71.3}   & 67.1 \\ \hline \hline
ChemProt          & \multicolumn{1}{c|}{15.9}      & \multicolumn{1}{c|}{68.9}   & 25.9 & \multicolumn{1}{c|}{17.9}      & \multicolumn{1}{c|}{62.0}   & 27.7 & \multicolumn{1}{c|}{22.0}      & \multicolumn{1}{c|}{69.7}   & 33.4 & \multicolumn{1}{c|}{19.0}      & \multicolumn{1}{c|}{60.6}   & 28.7 \\ \hline
DDI               & \multicolumn{1}{c|}{99.6}      & \multicolumn{1}{c|}{48.6}   & 16.1 & \multicolumn{1}{c|}{19.9}      & \multicolumn{1}{c|}{79.1}   & 31.8 & \multicolumn{1}{c|}{25.5}      & \multicolumn{1}{c|}{77.9}   & 38.4 & \multicolumn{1}{c|}{17.3}      & \multicolumn{1}{c|}{75.4}   & 28.2 \\ \hline
GAD               & \multicolumn{1}{c|}{51.4}      & \multicolumn{1}{c|}{92.3}   & 66.0 & \multicolumn{1}{c|}{63.7}      & \multicolumn{1}{c|}{57.2}   & 60.2 & \multicolumn{1}{c|}{64.1}      & \multicolumn{1}{c|}{78.5}   & 70.3 & \multicolumn{1}{c|}{63.2}      & \multicolumn{1}{c|}{72.7}   & 67.6 \\ \hline
\textbf{RE Average} & \multicolumn{1}{c|}{25.6}      & \multicolumn{1}{c|}{70.0}   & 36.0 & \multicolumn{1}{c|}{33.8}      & \multicolumn{1}{c|}{66.1}   & 39.9 & \multicolumn{1}{c|}{37.2}      & \multicolumn{1}{c|}{75.4}   & 47.4 & \multicolumn{1}{c|}{33.2}      & \multicolumn{1}{c|}{69.6}   & 41.5 \\ \hline
\end{tabular}
\end{table*}

\section{Datasets Details}
When assessing biological information extraction, we employ the identical NER and RE datasets like the ones' which are employed in the BLURB benchmark \cite{gu2021domain}. The datasets' statistics are detailed in Table \ref{tab:Table 0}, along with other information. As stated in Section 2.3 of \cite{gu2021domain}, the processing and train/dev/test splits are carried out. As the premise of this paper revolves around the concept from \cite{gutierrez2022thinking}', we conducted a simulation on the dataset procured from the same source

\begin{enumerate}

    \item \textbf{NCBI-disease :} NCBI disease corpus\cite{dougan2014ncbi} a resource for disease name recognition and concept normalization. Journal of biomedical informatics, 47, 1-10.

    \item \textbf{BC2GM :} The Biocreative II Gene Mention corpus \cite{smith2008overview} is a dataset that includes 17,500 words with annotations for gene entities from PubMed articles. Recognizing the gene and protein described in text is the dataset's aim. It is frequently employed in tasks involving named entity recognition in biomedicine.

    \item \textbf{JNLPBA :} The Joint Workshop on Natural Language Processing in Biomedicine and its Applications dataset \cite{collier2004introduction}is composed of 2,000 MEDLINE abstracts which have been manually selected and annotated for entities related to genes.

    \item \textbf{BC5CDR : } The BioCreative V Chemical-Disease Relation corpus\cite{li2016biocreative}, which contains PubMed abstracts with annotations of both diseases and chemicals, is used for evaluation of models in the biomedical information extraction task. The models are evaluated separately for each entity type, following the approach used in previous studies \cite{gu2021domain}.

    \item \textbf{ChemProt : }The ChemProt dataset \cite{krallinger2017overview} comprises 1,820 abstracts from PubMed that are labeled with annotations for chemical-protein interactions. The dataset has six relation categories, five of which are true and one is considered vacuous.

    \item \textbf{DDI : }The DDI dataset, as described by \cite{herrero2013ddi}, includes sentences from MEDLINE and DrugBank that are labeled with information about drug-drug interactions. The interactions are categorized into four true relations and one vacuous relation.

    \item \textbf{GAD : }The Genetic Association Database corpus\cite{bravo2015extraction} is a collection of scientific texts, including excerpts and abstracts, that have been annotated with gene-disease associations in a distant manner, meaning that the annotations were made based on information inferred from the text rather than directly stated in it.
    
\end{enumerate}

\begin{table}
\centering
\caption{\label{tab:Table 0}Dataset details and statistics}
\begin{tabular}{|l|l|l|l|l|l|}
\hline
               & Task & Train & Dev   & Test  & Eval. Metric    \\ \hline
BC5CDR-disease & NER  & 4182  & 4244  & 4424  & F1 entity-level \\ \hline
BC5CDR-chem    & NER  & 5203  & 5347  & 5385  & F1 entity-level \\ \hline
NCBI-disease   & NER  & 5134  & 787   & 960   & F1 entity-level \\ \hline
JNLPBA         & NER  & 46750 & 4551  & 8662  & F1 entity-level \\ \hline
BC2GM          & NER  & 15197 & 3061  & 6325  & F1 entity-level \\ \hline \hline
DDI            & RE   & 25296 & 2496  & 5716  & Micro F1        \\ \hline
ChemProt       & RE   & 18035 & 11268 & 15745 & Micro F1        \\ \hline
GAD            & RE   & 4261  & 535   & 534   & Micro F1        \\ \hline
\end{tabular}
\end{table}

\subsection{Analyzing comparable methods}
Using 100 training instances, the researchers ran tests to compare the effectiveness of four pre-trained language models: PubMedBERT-base\cite{gu2021domain}, BioBERT-large\cite{lee2020biobert}, RoBERTa-large\cite{liu2019roberta}, and GPT-3. RoBERTa-large was previously trained on general-domain text, while PubMedBERT-base and BioBERT-large were pre-trained on a sizable collection of biomedical literature from PubMed. The in-context prompt for each test example was retrieved by the researchers for the GPT-3 by employing the identical hundred training sets. This observation has been derived from the research conducted in \cite{gutierrez2022thinking}.
\\\textbf{Implementation Details.}We chose to use 100 annotated examples for their experiments, as they considered this a reasonable number to start training an information extraction model for a new assignment. We used a balanced set of 100 examples for the relation extraction task, with an equal distribution across relation types. From the previous work, we found that the BERT-sized pre-trained language models were fine-tuned using the HuggingFace Transformers library\cite{wolf2019huggingface}, while for the GPT-3 experiments, we used a maximum of 10 and 5 in-context examples for NER and RE, respectively, to stay within GPT-3's input length limit. Because GPT-3 is expensive, we only assessed each approach on a maximum of 1,000 test cases from each dataset, sampled stratified to match the distribution of relation types in the original test set. We chose models and prompt designs using the true few-shot framework, and carried out all experiments using three distinct training sets of 100 examples, reporting the mean and standard deviation to take training data volatility into account.

\section{Results and Analysis}

Table \ref{tab:comparison} contains the major findings of our experiment. As the same dataset was used for the stimulation, the resulting findings are comparable to those reported in \cite{gutierrez2022thinking}. Our findings demonstrate that, across all datasets, fine-tuned BERT-sized PLMs outperform GPT-3 in-context learning, frequently by a large margin (with an average improvement of 15.6-16.7\% for NER and 3.9-11.4\% for RE in F1 scores). As GPT-3's recall falls by twice as much in NER despite a 10-point decline in precision, this shows that GPT-3's under-prediction of entities plays a large role in its subpar in-context learning performance. On the other hand, GPT-3 performs poorly in the "none" relation class in RE tasks, which contributes to the sharper decline in precision.

We found that, despite the tiny size of the training sets, BERT-sized PLMs exhibit respectable performance in NER tasks after assessing the findings of our fine-tuning experiment. On account of the great lexical regularity of names of drugs, we specifically achieved strong results in the drug extraction task (BC5CDR-chem) with scores in the mid-80s. However, due to the higher lexical variability in the names of these entities, performance in other biomedical NER tasks, such as illness and gene extraction, declines to the high and low 60s. Recent research demonstrates that PLMs tuned on complete training sets also exhibit similar performance disparity. Furthermore, we found that the basic PubMedBERT model outperformed more complex variants of the general-domain RoBERTa and the biomedicine-specific BioBERT models, suggesting that pre-training on domain-specific text and vocabulary from beginning is particularly beneficial for NER tasks. These conclusions are in line with the findings of earlier research by \cite{gu2021domain}. 

Due to the assessed techniques' increased complexity, particularly in DDI and ChemProt, which have more relation types and class imbalance, the performance of these methods diminishes in RE tasks. RoBERTa-large outperforms PubMedBERT-base and BioBERT-large in the relation extraction task, in contrast to the NER task \cite{gu2021domain} and other studies with larger training sets. This suggests that, for activities requiring sophisticated syntactic and semantic comprehension, like RE, larger-scale general-domain pre-training can offset the advantages of domain-specific pre-training.

\subsection{GPT-3: Analyzing the effects of removing components}

We conducted ablation tests on a portion of 250 validation cases from sample datasets for every task, as shown in Tables \ref{tab:Table 2} and \ref{tab:Table 3}" in order to examine the methods employed to enhance GPT-3's efficiency. We  replaced the kNN module for another one that arbitrarily selects instances from the training set to be the in-context prompts for each test example. The findings demonstrated that the kNN module's removal decreased the GPT-3's in-context learning performance, with RE showing a more pronounced performance decline than NER.

We discovered that eliminating the logit bias option decreased recall even though precision improved in their NER-specific ablation trial. This was due to the post-processing step that removes predicted entities not present in the original sentence, leading to fewer false positives but also a decrease in the number of valid spans predicted. When the kNN module and logit bias option were both removed, the drop in performance was even greater, indicating that they complement each other.

With or without the kNN module, we discovered in the RE-specific excision investigation that eliminating the calibration module decreased precision and recall, demonstrating the efficiency of the module.

\begin{table}
    \centering
    \label{tab:Table 2}
    \caption{NER ablation study on BC5CDR-disease}
    \begin{tabular}{|l|l|l|l|}
\hline
            & Precision & F1   & Recall \\ \hline
Best Model  & 42.5      & 46.3 & 50.9   \\ \hline
Logit Biases & 66.7      & 42.6 & 31.3   \\ \hline
kNN Module  & 42.7      & 46.3 & 50.9   \\ \hline
Both        & 60.2      & 38.7 & 28.5   \\ \hline

    \end{tabular}
\end{table}

\begin{table}
    \centering
    \caption{RE ablation study on DDI.}
    \label{tab:Table 3}
    \begin{tabular}
{|l|l|l|l|}
\hline
            & Precision & F1   & Recall \\ \hline
Best Model  & 16.1      & 26.1 & 68.0   \\ \hline
Calibration & 14.6      & 23.6 & 62.0   \\ \hline
kNN Module  & 11.5      & 18.6 & 48.0   \\ \hline
Both        & 10.9      & 16.9 & 38.0   \\ \hline
    \end{tabular}
\end{table}

\subsection{Error Analysis}

In this section, we conducted a thorough analysis and discovered that in-context learning struggles with handling the null class. This refers to sentences that do not contain any entities (for NER) and entity pairs that do not have any of the target relations (for RE). We found that these difficulties are not unique to biomedical applications and are likely to be problematic for Information Extraction (IE) tasks in general.

\begin{table}
    \centering
    \caption{Assessment on modified BC5CDR-disease where sentences without disease entity are not removed.}
    \label{tab:Table 4}
    \begin{tabular}
{|llll|}
\hline
\multicolumn{4}{|l|}{\textbf{Original BC5CDR-disease}}                                                                \\ \hline
\multicolumn{1}{|l|}{}                 & \multicolumn{1}{l|}{Precision} & \multicolumn{1}{l|}{F1}   & Recall \\ \hline
\multicolumn{1}{|l|}{RoBERTa-large}    & \multicolumn{1}{l|}{66.9}      & \multicolumn{1}{l|}{67.7} & 68.7   \\ \hline
\multicolumn{1}{|l|}{GPT-3 In-Context} & \multicolumn{1}{l|}{57.9}      & \multicolumn{1}{l|}{43.6} & 35.0   \\ \hline
\multicolumn{4}{|l|}{\textbf{Modified BC5CDR-disease}}                                                                \\ \hline
\multicolumn{1}{|l|}{}                 & \multicolumn{1}{l|}{Precision} & \multicolumn{1}{l|}{F1}   & Recall \\ \hline
\multicolumn{1}{|l|}{RoBERTa-large}    & \multicolumn{1}{l|}{68.0}      & \multicolumn{1}{l|}{70.4} & 72.9   \\ \hline
\multicolumn{1}{|l|}{GPT-3 In-Context} & \multicolumn{1}{l|}{60.3}      & \multicolumn{1}{l|}{59.8} & 59.3   \\ \hline
    \end{tabular}
\end{table}

\subsection{RE Error analysis}

Our study also looked at the impact of the null class (called "none relation" in the DDI dataset) in the RE task. Table \ref{tab:comparison} shows that when multiple relation types exist in the dataset, such as in DDI and ChemProt, GPT-3 in-context learning achieves high recall but low precision. Upon examining the confusion matrices generated through LOOCV, we discovered that GPT-3 rarely predicts the "none relation" in the DDI dataset. This bias against the "none relation" significantly lowers the model's precision, especially since the DDI dataset is heavily weighted towards this class.

\begin{table}
\centering
    \caption{Comparison of two DDI (drug-drug interaction) examples predicted by GPT-3 in-context learning and RoBERTa-large.}
    \label{tab:Table 6}
\begin{tabular}{|c|c|c|}
\hline
Label                   & Model         & Sample                                                                                                                                                                                                                   \\ \hline
\multirow{2}{*}{Effect} & RoBERTa-large & \begin{tabular}[c]{@{}c@{}}Concurrent use of \\ phenothiazines may\\ antagonize the \\ anorectic effect of \\ diethylpropion\end{tabular}                                                                                \\ \cline{2-3} 
                        & GPT-3         & \begin{tabular}[c]{@{}c@{}}Concurrent use of \\ phenothiazines may \\ antagonize the \\ anorectic effect of \\ diethylpropion.\end{tabular}                                                                              \\ \hline
\multirow{2}{*}{None}   & RoBERTa-large & \begin{tabular}[c]{@{}c@{}}Other powerful CYP3A4 \\ inhibitors (e.g., itraconazole,\\  clarithromycin, nefazodone, \\ troleandomycin, ritonavir, \\ nelfinavir) should \\ function similarly.\end{tabular}               \\ \cline{2-3} 
                        & GPT-3         & \begin{tabular}[c]{@{}c@{}}Other powerful CYP3A4 \\ inhibitors (e.g., itraconazole, \\ clarithromycin, nefazodone, \\ troleandomycin, ritonavir, \\ nelfinavir) are not\\  predicted to function similarly.\end{tabular} \\ \hline
\end{tabular}
\end{table}

In table \ref{tab:Table 6}, the evaluation of the comparison of LIME-based saliency scores for two DDI examples predicted by GPT-3 in-context learning and RoBERTa-large. involves masking out words highlighted in blue and observing the change in the model's current prediction. The drugs shown in bold are the head and tail entities for the relation being queried. The second example highlights that GPT-3 in-context learning is more susceptible to spurious surface-level signals, which can lead to incorrect predictions, particularly in predicting the none-class. To gain a better understanding of this bias, we used LIME \cite{robertson2009probabilistic} to analyze the predictions made by both GPT-3 and RoBERTa on effect and none examples. The first example in Table \ref{tab:Table 6} was correctly labeled by both models by relying on the relevant signal of the "anorectic effect". However, correct predictions for none examples often require a more implicit understanding of the sentence's structure rather than relying on surface-level signals, as demonstrated in the second example in \ref{tab:Table 6}. Here, we noticed that RoBERTa's prediction is strongly influenced by the phrase "of CYP3A4 (e.g.," which suggests that the drugs within the parentheses belong to the same class and, therefore, do not interact with each other. This indicates that RoBERTa correctly uses the linguistic structure of the sentence. On the other hand, GPT-3's incorrect mechanism prediction seems to be supported by the phrase "expected to behave similarly," which is irrelevant to the relation being queried between the drugs. This suggests that GPT-3's in-context learning is more susceptible to spurious surface-level signals and, therefore, struggles with predicting the none class.

\subsubsection{NER Error Analysis}

When using a NER model in real-world scenarios, there may be many instances where the input sentence contains no relevant entity, which is referred to as a null class example. This is common in datasets like BC5CDR-disease, where up to 50\% of sentences contain no disease. However, previous studies on GPT-3 in-context learning have not taken this into account. For example, \cite{zhao2021calibrate} removed all examples that did not contain relevant slots from their slot-filling experiment. This approach ignores the impact of null class examples on the poor performance of in-context learning. Our study shows that null class examples significantly cause this issue.

Using a modified BC5CDR-disease dataset, where all words lacking disease entities were eliminated, we ran an experiment comparing GPT-3 in-context learning with fine-tuned RoBERTa-large to examine the effects of null samples. Table \ref{tab:Table 4} shows our results, which reveal that GPT-3's recall increased by about 24\% compared to only 4\% for RoBERTa-large, demonstrating that prompts having null examples strongly biases GPT-3 to predict few entities rather than contributing them to the fine-tuning data.

We believe that GPT-3's bias in underpredicting entities is partially due to the fact that in-context learning requires it to predict relevant entities only if they are present in the given sentence, which is different from how smaller PLMs predict entities. To test this hypothesis, we simplified our experiment by removing the k-NN retrieval module and using the same two-shot prompt with one example having no entities and another with at least one entity for all examples in the BC5CDR-disease training dataset. We then added a random example without entities to every prompt and compared the probability of a null prediction in each setting. We found that adding the second null example slightly increased the null probability more for examples without entities than those with entities, accounting for the lower initial null probability for examples with one or more entities reversed this effect. The fact that the increase in null probability was not significantly larger for examples without entities suggests that GPT-3 struggles to infer the appropriate prediction constraint for this task and instead increases the null probability somewhat uniformly across examples. This is shown in Table \ref{tab:Table 5}. In table \ref{tab:Table 5}, We compare the null token probability assigned by GPT-3 to examples with zero and non-zero entities in the dataset using 2-shot and 3-shot prompts. The 3-shot prompts contain an additional null example to examine its effect. The evaluation presents the average over 3 randomly chosen prompts.

\begin{table}
    \centering
    \caption{Evaluation of GPT-3's performance on the BC5CDR-disease training dataset.}
    \label{tab:Table 5}
    \begin{tabular}
    {|l|l|l|l|l|}
\hline
Entity Number &
  \begin{tabular}[c]{@{}l@{}}P(null)\\ 2-Shot\end{tabular} &
  \begin{tabular}[c]{@{}l@{}}P(null)\\ 3-Shot\end{tabular} &
  \begin{tabular}[c]{@{}l@{}}Absolute\\ 
   $\Delta$\end{tabular} &
  \begin{tabular}[c]{@{}l@{}}\%\\ Increase\end{tabular} \\ \hline
One or More &
  15.8 &
  40.9 &
  25.1 &
  159\% \\ \hline
Zero(null) &
  19.4 &
  49.1 &
  29.7 &
  153\% \\ \hline
    \end{tabular}
\end{table}

\section{Implementation in Knowledge Base}

The Information Extracted by GPT-3 using the proposed approach is now applied in Knowledge Base Construction. For that, the following procedure should be followed:- 

\begin{enumerate}

    \item \textbf{Defining the domain and scope of the knowledge base}: To use prompt-based language models for Knowledge Base Construction, the knowledge base's domain, and scope must first be defined. This will make it easier to decide what kinds of entities and relationships should be recorded in the knowledge base. 

    It is a crucial step in implementing GPT-3 for information extraction in a knowledge-based system like a knowledge base construction (KBC) system. This step involves specifying the subject matter that the knowledge base will cover and the specific types of information that the system will extract and store.

    The domain of the knowledge base refers to the specific area of knowledge or expertise that the system will focus on. For example, the domain could be healthcare, finance, law, or any other field that requires specialized knowledge. Defining the domain is important because it will determine the type of information that the system will extract and store.

    The scope of the knowledge base refers to the breadth and depth of the information that the system will cover. For example, the scope could be limited to a specific area within the domain, such as medical diagnosis or financial planning, or it could be broader, covering multiple areas within the domain. Defining the scope is important because it will determine the level of detail that the system will extract and store.

    Once the domain and scope of the knowledge base have been defined, the system can be designed to extract and store relevant information using GPT-3. GPT-3 can be used to analyze text and extract key information, such as named entities, relationships between entities, and other relevant data. The extracted information can then be stored in a structured format within the knowledge base, making it easily accessible and usable for a wide range of applications.
    
    \item \textbf{Preparing prompts}: The next phase is preparing the prompts to produce text that may be used for information extraction.Preparing prompts is an important step in implementing GPT-3 for information extraction in a knowledge-based system like a knowledge base construction (KBC) system. Prompts are the inputs given to GPT-3 to generate outputs based on the specific task at hand. The prompts define the context and requirements for the information extraction task and guide the model to generate relevant responses.

    To prepare prompts for GPT-3, you will first need to define the specific types of information that you want the system to extract. This could include entity types, relationships between entities, specific attributes or properties of entities, or any other relevant information. For example, if you are building a knowledge base on medical conditions, you may want the system to extract information such as symptoms, causes, treatments, and risk factors for each condition.

    Once you have defined the information to be extracted, you can create prompts that provide context for GPT-3 to generate responses. Prompts can be simple or complex, depending on the specific requirements of the task. For example, a simple prompt might ask GPT-3 to identify the symptoms of a particular medical condition, while a more complex prompt might ask the system to identify the symptoms, causes, and treatments of a condition and their interrelationships.

    To ensure that GPT-3 generates accurate and relevant responses, it is important to fine-tune the prompts by testing them with sample inputs and refining them based on the model's outputs. This process of trial and error can help improve the accuracy and relevance of the information extracted by GPT-3.

    Overall, preparing prompts is a critical step in implementing GPT-3 for information extraction in a KBC system, as it defines the context and requirements for the information extraction task and guides the model to generate relevant and accurate responses.

    \item \textbf{Collecting unstructured data}:Collecting unstructured data is a crucial step in implementing GPT-3 for information extraction in a knowledge-based system like a knowledge base construction (KBC) system. Unstructured data refers to data that is not organized in a structured format, such as text data in emails, social media posts, news articles, and other sources. Collecting unstructured data involves identifying relevant sources of information and extracting the text data from those sources.

    There are several ways to collect unstructured data, including web scraping, APIs, and manual collection. Web scraping involves automatically extracting data from web pages using tools like web crawlers or scraping software. APIs (Application Programming Interfaces) allow developers to access data from online services such as Twitter or Facebook programmatically. The manual collection involves manually copying and pasting data from sources like news articles or emails.

    Once the unstructured data has been collected, GPT-3 can be used to analyze the text and extract relevant information. GPT-3 is a language model that has been trained on a large corpus of text data and can generate high-quality responses to a wide range of natural language processing tasks. It can be used to identify named entities, relationships between entities, and other relevant information from unstructured text data.

    After the relevant information has been extracted from the unstructured data, it can be stored in a structured format within the knowledge base. This structured data can then be used to answer questions, generate insights, and provide recommendations based on the knowledge stored within the system.
    
    \item \textbf{Extracting structured data}: In implementing GPT-3 for information extraction in a knowledge-based system like a knowledge base construction (KBC) system, extracting structured data involves converting unstructured text data into a structured format that can be easily stored and analyzed by the system.

    GPT-3 can be used to analyze text and extract key information, such as named entities, relationships between entities, and other relevant data. Once the information has been extracted, it can be structured into a standardized format, such as a table or a graph, that can be easily stored and analyzed.

    There are several techniques that can be used to extract structured data from unstructured text using GPT-3. These include: 

    \begin{itemize}
    
        \item Named entity recognition (NER): This technique involves identifying and extracting entities such as people, places, organizations, and other types of named entities from unstructured text.
        
        \item Relationship extraction: This technique involves identifying and extracting relationships between entities in unstructured text. For example, if the text mentions a person and a company, the system can extract the relationship between the person and the company.

        \item Semantic parsing: This technique involves analyzing the syntax and semantics of a sentence or paragraph to extract structured information. For example, if the text mentions a price, a date, and a product name, the system can extract the structured data into a table format.
        
    \end{itemize}

    Once the structured data has been extracted, it can be stored in a database or knowledge graph and used for a wide range of applications, such as natural language understanding, question answering, and information retrieval.
    
    \item \textbf{Verifying and updating data}: To assure correctness and completeness, the retrieved data should be reviewed and updated as needed. Manual or automatic verification methods could be used.

    Verifying and updating data is an important part of information extraction in a knowledge-based system that uses GPT-3. After the system has extracted information using GPT-3, it needs to verify the accuracy of the extracted information and update it if necessary.

    Verifying data involves comparing the extracted information with existing data in the knowledge base to ensure that it is accurate and consistent. This can be done using various techniques, such as comparing named entities, checking relationships between entities, and cross-referencing with external sources. If the extracted information is found to be accurate, it can be stored in the knowledge base. If there are discrepancies or inconsistencies, the system can flag the information for further review or update it accordingly.

    Updating data involves modifying existing information in the knowledge base based on new information that has been extracted using GPT-3. This can happen when new information becomes available or when existing information needs to be corrected or updated. The system can use GPT-3 to extract new information and compare it with existing data in the knowledge base. If the new information is found to be accurate, the system can update the relevant entries in the knowledge base accordingly.

    Overall, verifying and updating data is an iterative process that involves continuously extracting new information and comparing it with existing data in the knowledge base to ensure accuracy and consistency. This helps to ensure that the knowledge base remains up-to-date and reliable, which is essential for its usefulness in knowledge-based systems like KBC.

    \item \textbf{Use of the knowledge base}: Once GPT-3 has been implemented for information extraction in a knowledge-based system like a knowledge base construction (KBC) system, the KBC system can be used for a wide range of applications, such as: 

    \begin{itemize}
    
        \item Answering questions: The KBC system can be used to answer questions related to the domain and scope of the knowledge base. Users can input a question and the system will search through the knowledge base to find relevant information and provide an answer.

        \item Decision-making: The KBC system can be used to support decision-making by providing relevant information and insights. For example, in healthcare, the system can provide recommendations for diagnosis and treatment based on the patient's symptoms and medical history.

        \item Research and analysis: The KBC system can be used to analyze data and provide insights that can be used for research and analysis. For example, in finance, the system can analyze financial data and provide insights into market trends and investment opportunities.

        \item Training and education: The KBC system can be used for training and education by providing access to a wealth of knowledge and information in the domain. For example, in law, the system can provide access to case law and legal precedents, which can be used for training and education purposes.

    \end{itemize}

    Overall, the KBC system can support a wide range of applications and provide valuable insights and information in the domain. By leveraging GPT-3 for information extraction, the KBC system can provide accurate and relevant information, making it a valuable tool for a wide range of users, including professionals, researchers, and students.

\end{enumerate}

\section{Conclusion}

In this study, we looked into how GPT-3 in-context learning may be used for the crucial job of information extraction (IE). Considering that such a paradigm would offer important benefits for biomedical IE applications, we invested a lot of time investigating the methods that have been successfully used in other in-context learning contexts. We demonstrated, however, that using a variety of benchmark datasets for biomedical NER and RE, existing methods do not allow GPT-3 in-context learning to outperform BERT-sized PLM fine-tuning. Also, we spoke about a few potential general restrictions on in-context learning in biomedical IE that will be investigated in subsequent research, including its difficulties in managing the null class, such as entity-less NER examples and vacuous relation examples for RE. In addition to presenting this topic for additional research, we expect that our work will be able to point biomedical researchers and practitioners in the direction of more effective and affordable methods for low-resource IE, including tiny PLM fine-tuning or possibly even directly fine-tuning GPT-3.

One of the primary benefits of using GPT-3 for extracting information for Knowledge Base Construction (KBC) is its ability to process natural language text in a more human-like way. GPT-3 can understand the context and meaning of text and can extract relevant information, such as entities, relationships, and attributes, from unstructured data sources, such as news articles, social media posts, and web pages. This allows for the construction of a more sophisticated and intelligent knowledge base that can be used to answer complex queries and perform advanced analysis. Additionally, GPT-3 can significantly improve the efficiency and accuracy of IE, reducing the need for manual curation and increasing the speed at which new information can be added to the knowledge base. Overall, the use of GPT-3 for IE in KBC can lead to a more intelligent and effective system that can provide more value to users and organizations.

\bibliographystyle{IEEEtran}
\bibliography{ref.bib}

\end{document}